\documentclass[conference]{IEEEtran}
\IEEEoverridecommandlockouts
\usepackage{cite}
\usepackage{url}
\usepackage{amsmath,amssymb,amsfonts}
\usepackage{algorithmic}
\usepackage{graphicx}
\usepackage{textcomp}
\usepackage[numbers]{natbib}
\usepackage{xcolor}
\def\BibTeX{{\rm B\kern-.05em{\sc i\kern-.025em b}\kern-.08em
    T\kern-.1667em\lower.7ex\hbox{E}\kern-.125emX}}
\begin{document}

\title{Semantic Computing for Organizational Effectiveness: From Organization Theory to Practice through Semantics-Based Modelling}

\author{\IEEEauthorblockN{Mena Rizk, Daniela Rosu, and Mark Fox}
\IEEEauthorblockA{\textit{Enterprise Integration Laboratory} \\
\textit{University of Toronto}\\
Toronto, Canada \\
mena.rizk@mail.utoronto.ca, drosu@mie.utoronto.ca, msf@eil.utoronto.ca}

}


\maketitle

\begin{abstract}
A critical function of an organization is to foster the level of integration (coordination and cooperation) necessary to achieve its objectives. The need to coordinate and motivation to cooperate emerges from the myriad dependencies between an organization’s members and their work. Therefore, to reason about solutions to coordination and cooperation problems requires a robust representation that includes the underlying dependencies. We find that such a representation remains missing from formal organizational models, and we leverage semantics to bridge this gap. Drawing on well-established organizational research and our extensive fieldwork with one of North America's largest municipalities, (1) we introduce an ontology, formalized in first-order logic, that operationalizes concepts like outcome, reward, and epistemic dependence, and their links to potential integration risks; and (2) present real-world applications of this ontology to analyze and support integration in complex government infrastructure projects. Our ontology is implemented and validated in both Z3 and OWL. Key features of our model include inferable dependencies, explainable coordination and cooperation risks, and actionable insights on how dependency structures within an organization can be altered to mitigate the risks. Conceptualizing real-world challenges like incentive misalignment, free-riding, and subgoal optimization in terms of dependency structures, our semantics-based approach represents a novel method for modelling and enhancing coordination and cooperation. Integrated within a decision-support system, our model may serve as an impactful aid for organizational design and effectiveness. More broadly, our approach underscores the transformative potential of semantics in deriving tangible, real-world value from existing organization theory.
\end{abstract}


\section{Introduction}
Effective coordination and cooperation are at the heart of high-performing organizations. These capabilities both shape and are shaped by the web of dependencies between organizational members and their tasks \cite{2020_raveendran.silvestri.ea_RoleInterdependenceMicroFoundationsOrganizationDesign}.  Therefore, to improve coordination and cooperation in organizations, we must first have a robust representation of the dependencies within them. In this paper, we present a novel formal representation of dependencies, designed to support organizations in overcoming coordination and cooperation challenges. Our research is grounded in a long-term collaboration with one of North America's ten most populous cities. 

In Section 2 we present a selection of coordination and cooperation challenges faced during a large-scale infrastructure project to motivate and discuss the requirements for a semantics-based solution geared towards their resolution. In Section 3, we delve into relevant organizational literature, highlighting the link between epistemic, outcome, and reward dependencies and an organization's coordination and cooperation capabilities. We then present a survey of existing representational frameworks of dependence, coordination, and cooperation. 
In Section 4, we present our approach to conceptualizing 
epistemic, outcome, and reward dependencies, and the pathways that lead to coordination and cooperation risks. Formalization using first-order logic, solvability, and computational implementation is discussed in Section 5. In Section 6 we present a real-world application to a use case provided by our collaborators. We discuss in Section 7 the strengths and limitations of our formalization, as well as our future work plans. In Section 8 we highlight broader implications beyond organizational (re)structuring. 

In sum, the contributions of this paper are focused on leveraging semantic technologies to assist with putting organization theory into practice and are three-fold: 1. a semantics-based methodology for addressing real-world coordination and cooperation challenges; 2. a formal conceptual model of dependency, coordination, and cooperation, expressed as an ontology and 3. a proof-of-concept semantic decision support system anchored by the conceptual model.  

\section{Motivation}
Coordination across divisions, stakeholders, projects, and resources is essential for successful project delivery. Current approaches, which place heavy reliance on human intellect and manual effort, are not only susceptible to errors but also difficult to expand and implement on a larger scale. These approaches lean heavily on human reasoning and intervention, rendering them less efficient and more mistake-prone than alternatives that leverage information technology.

As organizations and projects expand in size and complexity, facilitating efficient coordination becomes progressively more challenging, and this challenge extends to government entities as well. Our research partner, the municipal government (“the City”) of one of the ten largest cities in North America, is facing difficulties in managing multiple complex infrastructure projects simultaneously. Our partnership aims to investigate how information technology, and decision support systems, in particular, can help the City enhance its coordination of such complex projects.

In this section we offer a concise overview of one such undertaking, providing insights into how advanced technologies can enhance municipal decision-making processes. The endeavour in question focuses on the City's initiatives to reduce flood risk in one of its most vulnerable areas. The Riverine Flooding Project (RFP), aimed at reducing riverine flooding risks in the neighbourhood, involves designing and implementing several pieces of infrastructure such as bridges and concrete channels. Concurrent with this, the City’s Urban Flooding Program (UFP) is implementing sewage infrastructure improvements that must be coordinated with the RFP. There are also several planned capital works projects in the area that the RFP must consider in its design. 

As operating in this complex environment entails intense coordination across various organizations, stakeholders, and tasks, we had access to a wealth of coordination phenomena and challenges to model and help address. A detailed report on the case studies we participated in will be released separately, and we provide here a brief summary. 

We identified three types of coordination challenges: participation, navigation, and cross-cutting collaboration. \textit{Participation} relates to delays caused by inadequate stakeholder responsiveness. \textit{Navigation} involves difficulties in identifying affected stakeholders and accounting for their needs. \textit{Cross-cutting collaboration} is about the unsynchronized efforts of numerous stakeholders across bureaucratic silos to meet interdependent objectives, overlooking the broader project context. \textit{Failures} in participation, navigation, and cross-cutting collaboration lead to significant delays, cost overruns, and suboptimal implementations that require the rebuilding of various pieces of infrastructure.

The key issues underlying these challenges are deficiencies in a city's capacity to efficiently coordinate and cooperate. These deficiencies have roots in a lack of ability to recognize the dependencies that necessitate coordination and cooperation and the reasons for their failures. Therefore, an effective decision-making aid should be able to help with recognizing the interdependencies that require coordination, as well as the factors that prevent cooperation from occurring when needed and assist in developing appropriate strategies to mitigate the risk of future failures. Automatically identifying instances of dependence, and the areas that need coordination and those currently experiencing failures require grounding in a robust representation of dependence, the entities through which dependencies materialize, and the factors leading to failures. 

Though much effort has been devoted in the organizational literature to conceptualizing and addressing these challenges, existing representational frameworks are limited in their ability to model dependencies, identify coordination and cooperation risks, and be used as a tool for reasoning about alternative interventions for mitigating said risks (a more detailed discussion is provided in Section 3).

We, therefore, found it necessary to develop a formalization of organizational dependencies that supports the automated inference of coordination needs and cooperation risks, allowing organizations to improve their ability to detect risks of failure and determine how they can reduce these risks through different organizational designs that modify dependency patterns.

The inherent explainability of the inferences provided by knowledge-based systems not only allows for the identification of failure risk pathways, but also permits the direct simulation and evaluation of different modes of intervention, and alternative organizational and workflow design patterns.  In essence, with a causal model of dependence, coordination, and cooperation, organizations are afforded the ability to flag, explain, and confirm risks, as well as to estimate the effects of alternative solutions on the risks. 

\section{Related Work}
\subsection{Organizational Research}
Integration, defined by \citet{1967_lawrence.lorsch_DifferentiationIntegrationComplexOrganizations} as the collaborative harmony among departments essential to achieve unity of effort, rests on two pillars: coordination (alignment of actions) and cooperation (alignment of interests) \citep{2005_gulati.lawrence.ea_AdaptationVerticalRelationshipsIncentiveConflict,2017_kretschmer.vanneste_CollaborationStrategicAlliancesCooperationCoordination}. To integrate organizational efforts, it is critical to ensure agents possess necessary information (to facilitate coordination) and are motivated through appropriate incentives (to facilitate cooperation) \citep{2018_puranam_MicrostructureOrganizations}. Coordination failures arise primarily from knowledge gaps about others' actions. Similarly, cooperation failures stem from motivational discrepancies. If the reward does not commensurate with an agent's effort, or performance cannot be easily measured, the agent will exert themselves only to the extent they perceive necessary \citep{holmstrom1994firm,jensen1976theory}. These failures can manifest in various ways. Free-riding occurs when collective evaluations do not discern individual contributions or when individual monitoring costs are too high \citep{1984_jones_TaskVisibilityFreeRidingShirking}. Shirking occurs when an agent avoids duties, which may happen when monitoring is challenging, and consequences for low effort are perceived as minimal \citep{alchian1972production}. Sub-goal optimization arises when agents prioritize narrow individual objectives at the expense of broader organizational goals \citep{march1958organizations}.

The interplay between coordination and cooperation is significantly influenced by how agents depend on one another. Addressing challenges in coordination and cooperation necessitates understanding epistemic, outcome, and reward dependence. \citet{2012_puranam.raveendran.ea_OrganizationDesignEpistemicInterdependencePerspective} define epistemic dependence as instances where one agent's optimal action relies on predicting another agent's behaviour. It is the presence of epistemic dependence between two agents that generates the need to coordinate between them. Coordination failures can then be understood as stemming from unmet predictive needs between epistemically dependent agents. To preempt such risks, organizations can either establish proper coordination mechanisms to provide the information necessary to effectively align actions or reduce epistemic dependence.

\citet{1997_wageman.baker_IncentivesCooperationJointEffectsTask} describe reward dependence as the extent to which an individual’s rewards are influenced by coworkers' performance. Outcome dependence, on the other hand, pertains to whether an agent's work is appraised at an individual or group level \citep{2020_raveendran.silvestri.ea_RoleInterdependenceMicroFoundationsOrganizationDesign}. Reducing risks associated with cooperation failures often involves strategies aimed at increasing a sense of shared interests. One core method to ensure agents' interests align is the organization's formal incentive structure. \citet{2018_puranam_MicrostructureOrganizations} suggests that incentive structures vary along two axes: incentive \textit{breadth} (how much others' contributions factor into incentive decisions) and incentive \textit{depth} (how reliant incentives are on performance). When agents collectively share responsibility for evaluated and rewarded outcomes, their interests sync towards the shared results, thereby promoting cooperative behaviour. Reward and outcome dependence can be used to model the alignment of interests between agents, giving us insight into existing cooperation risks and how an organization designer may act to mitigate such risks. For instance, increasing incentive breadth and the level of aggregation at which the work of agents is measured and appraised corresponds to increasing the levels of reward and outcome dependence between agents, respectively.

\subsection{Related Semantic Frameworks from Enterprise Modelling}

We described in the previous section how analyzing dependencies offers a pathway to understanding the need for integration, the nature of integration challenges, and how the risk for such challenges can be mitigated. Bringing these insights to practical utility to support reasoning about how organizations can be structured to be better integrated, requires having a means of adequately representing the dependencies through which integration is necessitated and facilitated. In this section, we present a brief overview of extant frameworks for capturing the different forms of dependence and their relationship to coordination and cooperation in organizations. We focused on the enterprise modelling literature and examined the extent to which different frameworks support the representation of task, outcome, reward, and epistemic dependence in organizations. 

Enterprise modelling (EM), defined by Fox and Gr\"{u}ninger \cite{1998_fox.gruninger_EnterpriseModeling} as “a computational representation of the structure, activities, processes, information resources, people, behaviour, goals and constraints of a business, government, or other enterprise”, has generated various complementary frameworks over four decades. These are centred around activities (e.g., IDEF0 \cite{1998_menzel.mayer_IDEFFamilyLanguages}), business processes (e.g., CIMOSA \cite{cimosa}), and enterprise knowledge (e.g., TOVE \cite{1993_fox.chionglo.ea_COMMONSENSEMODELENTERPRISE}). While these frameworks incorporate elements of agents, goals, tasks, and activities, they do not offer a sufficiently robust means for representing dependencies of interest to our use cases.

As shown by Rizk et al. \cite{rizk2023}, there are two major limitations with most EM frameworks. Firstly, they do not distinguish between tasks and activities and therefore do not support modelling tasks as the intention to work towards a goal and activities as the precise ways in which one can act in order to achieve a goal. The inability to model this distinction limits the capacity of EM frameworks to represent ill-understood task environments where the precise actions necessary to achieve a goal cannot be expressed. Secondly, in the extant EM frameworks dependencies between tasks are either implicitly defined or asserted.  
The absence of an explicit mechanism for articulating their source and nature (which would allow for task dependencies to be inferred), circumscribes the ability to reason about how dependencies ought to be supported and managed (via coordination mechanisms). Given that a primary requirement that emerged from the problem domain is the ability to identify situations where coordination is needed in otherwise difficult-to-detect situations, being restricted to asserted dependencies largely defeats the point of applying semantic computing  for supporting coordination.

Existing EM frameworks were also found wanting in their ability to capture the dependencies between agents, owing to their limited or nonexistent support for modelling evaluative and incentivizing mechanisms that influence agent behaviour and performance. 
Besides their inability to capture epistemic, outcome, or reward dependence, these frameworks also offer scant provisions for modelling integration risks, which further circumscribes their utility in addressing the complex integration challenges identified in Section 2.

\section{Conceptual Model}

\begin{figure*}[!t]
\centering
\fbox{\includegraphics[width=16cm]{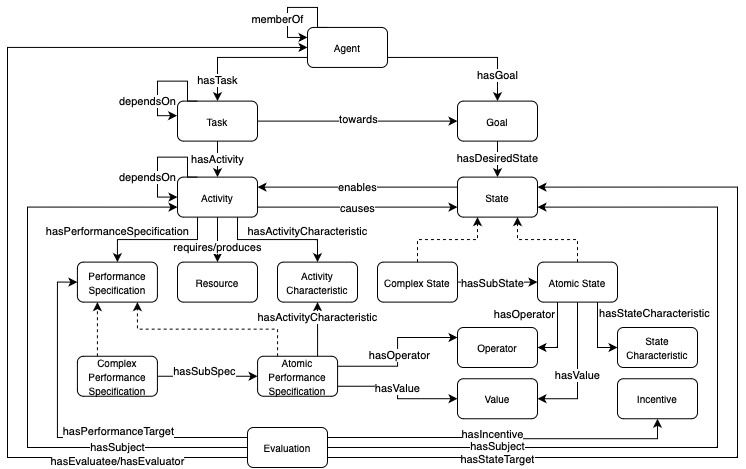}}
\caption{Simplified conceptual model. subClassOf relations are shown as dashed arrows. }\label{fig: odp}
\end{figure*}

In this section, we introduce our conceptual model (partially depicted in Figure 1), with each subsection covering a different layer. It is worth emphasizing that this model is not intended to exhaustively represent organizational structures, for which many frameworks exist \citep{2020_vernadat_EnterpriseModellingResearchReviewOutlooka}. Rather, it is intended to serve as a higher-level model of the different types of dependencies and their relationship to coordination needs and cooperation risks. 
Our framework is intended to help decision-makers capture the \textit{potential} for certain coordination needs and cooperation risks to occur as a consequence of dependency relationships. This conceptual model, and its formalization, can be extended with domain-specific theories for finer-grained reasoning, based on specialized field knowledge.
\subsection{Agents, Task Structure, and Task Dependence}
Central to our understanding of coordination and cooperation risks is the inherent task structure, which necessitates these practices. Within our framework, an {\fontfamily{phv}\selectfont Agent} is an entity capable of possessing goals, intending to strive towards them, and acting on those intentions. An agent can either be an {\fontfamily{phv}\selectfont Individual} or a {\fontfamily{phv}\selectfont Collective}, which in itself might be composed of individuals or other collectives. A {\fontfamily{phv}\selectfont Task} within our model denotes an agent's intention to contribute effort towards a goal. Conversely, a {\fontfamily{phv}\selectfont Goal} represents a desired ``state-of-affairs’’, defined in relation to the particular {\fontfamily{phv}\selectfont State} that is desired. Tasks may also be specified in terms of the activities through which goals can be achieved. The difference between tasks and activities is subtle but significant. While a task encapsulates the intention to act towards a goal, an activity specifies the precise way in which an agent can act. This distinction enables us to comprehend coordination and cooperation risks across a spectrum of task environments \citep{2020_raveendran.silvestri.ea_RoleInterdependenceMicroFoundationsOrganizationDesign} and addresses the problematic dual usage of the term ``task’’ as the achievement of goals and performance of activities in the literature \citep{1994_crowston_TaxonomyOrganizationalDependenciesCoordinationMechanisms,chandrasekaran1998ontology}.

An {\fontfamily{phv}\selectfont Activity} is a well-defined operation that an agent can perform as part of a task, causing an outcome state. It may also be enabled by a state and may require or produce a {\fontfamily{phv}\selectfont Resource}. An {\fontfamily{phv}\selectfont ActivityCharacteristic} represents a particular dimension of the performance of an activity that is subject to variation (such as start time, quality or design of output, and location of execution). A {\fontfamily{phv}\selectfont PerformanceSpecification} defines a particular specification of values for an activity’s characteristics. It can be either a {\fontfamily{phv}\selectfont ComplexPerformanceSpecification} or an {\fontfamily{phv}\selectfont AtomicPerformanceSpecification}, where the former is described in terms of the conjunction or disjunction of other complex or atomic specifications, and the latter is expressed in terms of a single {\fontfamily{phv}\selectfont ActivityCharacteristic}, {\fontfamily{phv}\selectfont Operator} (e.g., $\leq$, $\geq$, =, $\neq$)., and {\fontfamily{phv}\selectfont Value} combination.  The value represents the specific value of the characteristic defined in an atomic specification. It can be based on a unit of measure, an ordinal or nominal scale, or any data type. An agent is a {\fontfamily{phv}\selectfont contributorTo} an activity if they perform the activity and they are a {\fontfamily{phv}\selectfont soleContributorTo} the activity if no other agent is a contributor to it. A {\fontfamily{phv}\selectfont State} describes a particular aspect of an object or situation. Similarly to performance specification, It can be either a {\fontfamily{phv}\selectfont ComplexState} (which can be a conjunction or disjunction of other states) or an {\fontfamily{phv}\selectfont AtomicState}. An atomic state is defined in terms of a state characteristic, operator, and value combination. An agent is a {\fontfamily{phv}\selectfont contributorTo} a state if they perform an activity that causes that state and they are a {\fontfamily{phv}\selectfont soleContributorTo} the state if no other agent contributes to it.

A task {\fontfamily{phv}\selectfont dependsOn} another task if the way in which the goal of the former task is achieved is constrained by the way in which the goal of the latter is achieved. An activity {\fontfamily{phv}\selectfont dependsOn} another activity if the way in which the former is performed is constrained by the latter. Note that these relations hold irrespective of the agents that have the tasks or perform the activities. The efforts of agents may in some cases be strategic complements or substitutes. Two tasks or activities are {\fontfamily{phv}\selectfont strategicComplements} through a state if their combined effects on a state can be more or less than the sum of their individual contributions. Two tasks or activities are {\fontfamily{phv}\selectfont strategicSubstitutes} with respect to a state if they mutually offset the other (i.e. the value of one with respect to its effect on the state reduces the value of the other).

{\fontfamily{phv}\selectfont dependsOn}, {\fontfamily{phv}\selectfont strategicComplements}, and {\fontfamily{phv}\selectfont strategicSubstitutes} relationships can be directly asserted by experts, or inferred automatically based on sufficient conditions implemented by extensions with domain-specific knowledge.

\subsection{Evaluations, Incentives, and Predictive Needs}
Underpinning the notions of outcome, reward, and epistemic dependence are the concepts of evaluation, incentives, and predictive needs. An {\fontfamily{phv}\selectfont Evaluation} represents an instance of measurement or appraisal of the work of agents. It is defined in terms of an evaluatee(s) (the agent(s) whose work is being evaluated), evaluator(s) (the agent who is evaluating), the target of evaluation, and the subject of evaluation. The target of evaluation is the specific standard, benchmark, or expectation that the evaluatees are being measured against. A target can be a state or performance specification, which is akin to outcome and behaviour controls, respectively \citep{1979_ouchi_ConceptualFrameworkDesignOrganizationalControl}, allowing for a flexible representation of a wide variety of evaluation schemes in organizations. Each evaluation has exactly one target associated with it. The subject of evaluation is the particular efforts of an evaluatee which is being evaluated with respect to the target. Subjects can be either an activity performed by the evaluatee(s) or a state that is caused by the performance of their activity(s). Said differently, an evaluatee is answerable to an evaluator for the extent to which a target is satisfied, and the evaluation may consist of examining particular efforts of the evaluatee to assess their contribution to the target. An evaluation may also have an {\fontfamily{phv}\selectfont Incentive} associated with it. An incentive can be either a {\fontfamily{phv}\selectfont Reward} or a {\fontfamily{phv}\selectfont Sanction}. It is worth noting at this point that we only aim to capture “extrinsic” performance-based rewards and sanctions, such as performance bonuses and performance improvement plans. In this model, incentives are necessarily tied to evaluations since they are means for encouraging the contribution of effort towards performance or state targets, therefore these incentives are administered based on an appraisal of an evaluatee’s efforts. A {\fontfamily{phv}\selectfont predictiveNeed} is a quaternary relation between two agents and two tasks or activities. An agent has a predictive need of another agent if the former performs an activity (or has a task) that depends on the activity of the latter. Since the semantics of a dependency relation between two activities is that the way in which one activity is performed is constrained by another, the performer of the constrained activity must know how the constraining activity will be performed. 

\subsection{Agent Dependence, Coordination, and Cooperation}
We may now proceed with operationalizing the notions of outcome, reward, and epistemic dependence. An agent is {\fontfamily{phv}\selectfont outcomeDependentOn} on another agent through an evaluation if they are both evaluatees of that same evaluation. Two agents share in an evaluation when they are both individual evaluatees of the same evaluation or if they are both members of a collective which is itself the evaluatee of an evaluation. An agent is {\fontfamily{phv}\selectfont epistemicallyDependentOn} another agent through an evaluation if the evaluation has an incentive which the former agent is a recipient of, and the former agent has a predictive need of the latter agent due to an activity that is the subject of the evaluation. Given that the reward accrued by the former agent is contingent upon their performance of the activity that is the subject of evaluation, and the activity is constrained by another activity that is performed by the latter agent, the former agent is epistemically dependent on the latter since their reward depends on their ability to predict how the latter will perform their activity. There are two ways that an agent may be {\fontfamily{phv}\selectfont rewardDependentOn} on another agent through an evaluation. In the first case, if an agent is outcome-dependent on another agent through an evaluation that also has a reward that both agents are recipients of, then they are both also reward-dependent. Secondly, if an agent is epistemically dependent on another agent through an evaluation, then they are also reward-dependent on the same agent through the same evaluation. The second case holds by definition of reward dependence since the reward accrual of the former agent is dependent on the performance of the latter agent. It is worth reiterating that outcome, reward, and epistemic dependence are ternary relations consisting of two agents and an evaluation. The inclusion of an evaluation in these relations is imperative for delineating and tracing the particular instances of evaluations through which agents may depend on one another in different ways. 

The need to coordinate and the risks for cooperation failures between agents are captured based on the three types of dependencies described above. A {\fontfamily{phv}\selectfont coordinationNeed} exists between two agents when at least one is epistemically dependent on the other, since there is a requirement for at least one agent to have knowledge about the actions of the other to perform their work optimally. A {\fontfamily{phv}\selectfont cooperationRisk} exists between two agents  when there is a risk for free-riding, shirking, or sub-goal optimization between them. A {\fontfamily{phv}\selectfont freeRidingRisk} exists between an agent and an evaluation when an agent is an evaluatee of an evaluation that includes other evaluatees yet there is no subject of evaluation that the agent solely contributes to. This can lead to a cooperation risk between two agents if at least one agent in an evaluation has a free-ride risk with that evaluation. A {\fontfamily{phv}\selectfont shirkRisk} exists between an agent and a task (or activity) when an agent has a task or activity for which there is no evaluation. This poses a cooperation risk when one agent is epistemically dependent on another, yet there is a shirk risk between the latter agent and the work that the former agent’s work depends on. Finally, a {\fontfamily{phv}\selectfont subGoalOptimizationRisk} exists between two agents and a state when the state is part of some goal, both agents have tasks or activities that are strategic complements with respect to the state, there are evaluations for their individual work, yet there does not exist an evaluation for the complementary state. This may cause a cooperation risk since neither of the agents may be motivated to exert efforts towards the complementary state since they are only being evaluated on their individual contributions.

\section{Formalization and Implementation}
We have formalized our conceptual model in first-order logic and used it as the core of a proof-of-concept decision support system implemented in Python. Due to space limitations, we have made available the full formalization in a Github repository\footnote{\url{https://github.com/rizkmena/Organization-Dependence-Ontology/}}. First-order logic allows us to formalize relationships, such as epistemic dependence, in their intuitive format as predicates of higher arity, without having to resort to artificial modelling artifacts, as would be the case if restricting ourselves to syntactic fragments of first-order logic such as Description Logic. The formal representation constitutes the backbone of a decision support system that uses Z3\footnote{https://github.com/Z3Prover/z3}, a powerful external SMT-solver, as its main inference engine. The current formalization of the conceptual model does not make demands on the specialized algorithms for solving background theories, such as arithmetic, that SMT solvers come equipped with, however, extensions with domain-specific theories can take advantage of them, with the usual caveat that SMT solvers are not complete with respect to these theories. Using the basic theory (the formalization of the conceptual model) and problem-specific assertions, the SMT solver is able to construct models of cooperation and coordination scenarios that reveal both sources, as well as pathways to failure risks. We have tested our conceptual model, and its formalization by applying it to several use cases, one of which we detail in the next section. In addition to the Z3 implementation, the axiomatization was also implemented as a consistency-verified OWL ontology to support broader accessibility and experimentation by interested users. Both implementations are available in the aforementioned Github repository.

\section{Application}

\begin{figure*}
\centering
\fbox{\includegraphics[width=16cm]{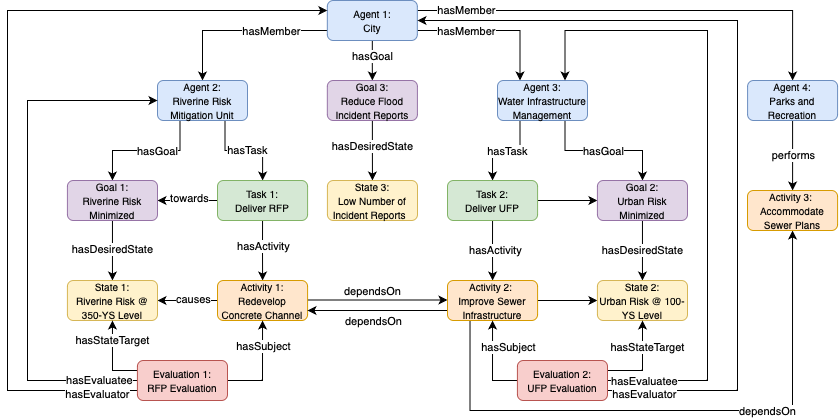}}
\caption{Partial representation of case study scenario assertions.}\label{fig: app}
\end{figure*}

To validate our framework's representational effectiveness and demonstrate its impact on supporting organizational integration we tested it on scenarios from our collaboration with the City. Given the space constraints, we discuss in this section an application to the project introduced in Section 2, focusing only on the essential scenario elements that allow us to illustrate how to capture dependencies and integration risks (see Figure 2). We begin with a description of the scenario which is then translated into model assertions, followed by a discussion of the model's inferences\footnote{The inferences claimed in our description have been verified in the Python/Z3 and Python/OWL implementations of the decision support system}. 

For the project of interest, we focus on three organizational units of the City: Riverine Flood Risk Mitigation (RM), Water Infrastructure Management (WIM), and Parks and Recreation (PR). The neighbourhood of interest is grappling with substantial flood risks due to both urban (sewer overflows) and riverine (watercourse overflows) sources. While RM is tasked with the implementation of the Riverine Flood Risk Mitigation Project (RFP) to combat riverine flood risks and aims to safeguard against a 350-year storm level through the construction of a concrete channel, WIM is simultaneously leading the Urban Flooding Program (UFP). This program aims to protect against a 100-year storm event by improving the sewer system. The City has consistently received numerous flooding incident reports and has the goal of reducing the total number of (reported) incidents.

Flooding is a complex issue due to riverine and urban flooding being interconnected sources. Enhancing the sewer system alone, for instance, while not addressing riverine flooding, still leaves the community exposed. Therefore, both the concrete channel development and sewer improvement activities are strategic complements with respect to the likelihood of a flooding incident occurring that would result in a report. More so, both activities depend on one another since the way in which either activity is designed and implemented imposes constraints on the other due to geographic overlap, funding constraints, and mutual impact on flood risk. Additionally, some sewer improvements must occur under recreational park space, which is under the jurisdiction of the PR unit. The PR unit performs the activity of reviewing and creating plans to accommodate the sewer improvement plans, which constrains WIM’s sewer improvement activities. RT and WIM are independently evaluated by the City on their activities for the achievement of their respective goal states.

Based on a formal encoding of this scenario using our framework, several coordination needs and cooperation risks were inferred. We briefly outline three:
\begin{itemize}
    \item \textit{Sub-goal optimization}: Since the activities of RT and WIM are strategic complements, yet both agents are only evaluated based on their individual targets, there is a potential cooperation risk between them since they have an incentive to prioritize that which they are directly answerable for, rather than the overall level of flood protection, which no one is evaluated on.
    \item \textit{Shirking}: To perform their activity of improving the sewer, WIM is epistemically dependent on PR since PR’s review and accommodation constrains the way in which sewer improvements can be made. However, since PR is not being evaluated on their review activity, they are not incentivized to accommodate WIM’s activities, especially because doing so would be costly for them. In essence, WIM needs PR but since PR will not need to answer to anyone for their efforts, there is a potential cooperation risk between them.
    \item \textit{Coordination risk}: Since the activities of both RT and WIM mutually constrain each other, and the incentives for both units are based on their respective activities, they are both epistemically dependent on one another, causing a coordination need between them. However, since no coordination mechanism was asserted to exist that would ensure they both have the necessary information to effectively execute their activities, there exists a potential coordination risk between them.
\end{itemize}

 These inferences allow the City to detect that a mitigation strategy, e.g., a coordination mechanism between RT and WIM, must be in place to ensure the successful integration of the concrete channel and the sewer improvements through the exchange of information necessary to manage their dependency. Additionally, the City is also made aware of the risk of cooperation failure between RT and WIM, which could be addressed by, for example, establishing evaluations and incentives to hold them collectively accountable for the overall flood risk in the neighborhood, rather than for their individual infrastructure delivery goals. Finally, by surfacing the cooperation risk between PR and WIM, the City is afforded the opportunity to set up a process for evaluating PR’s reviews that would promote a timely review and accommodation for the sewer improvements.

\section{Discussion and Future Work}
We draw attention to three key and mutually reinforcing benefits our representational framework offers for supporting organizational integration: inferable dependencies, explainable risks, and actionable insights into potential coordination and cooperation risks. Each of these benefits is afforded by the semantics-based axiomatization of a causal model for dependency-driven organizational integration analysis.

\begin{enumerate}
    \item \textbf{Inferable Dependencies}: The precise semantics of our axiomatization allows for the automated inference of epistemic, outcome, and reward dependencies based on the arrangement of work allocation, evaluations, and incentives in an organization. By inferring these types of dependencies, our model supports the detection of otherwise hidden relationships between agents, that can lead to integration failures, as we regularly observed in our case studies. Embedded within a decision-support tool, our model is able to preemptively flag dependencies, safeguarding projects, organizations, and communities from unforeseen complications.
    \item \textbf{Explainable Risks:} Predictions of risks can be backtracked to discern specific dependency relationships that instigate these risks. Such transparency allows organizational leaders to pinpoint the exact configurations of evaluations and incentives that contribute to cooperation risks. Oftentimes, simply making the challenges visible is enough to support leaders in reasoning about how to intervene. Our model elevates this approach by elucidating the underlying logic that explains why the network of risks exists in the first place.
    \item \textbf{Actionable Insights}: Beyond mere identification, our model fosters experimentation. With a clear view of risks, experts can simulate varying task allocations, incentive structures, and evaluation methods. This experimental facet enables the prediction of potential consequences, both intended and unintended, thereby refining intervention strategies.
\end{enumerate}

However, it's crucial to recognize the model's limitations. For one, it doesn't allow for differentiating between the quality or the significance of evaluations. As established by the organizational control literature, the observability and measurability of agent efforts play a pivotal role in influencing behaviour \citep{1979_ouchi_ConceptualFrameworkDesignOrganizationalControl,1985_eisenhardt_ControlOrganizationalEconomicApproaches} and without accounting for ``measurement quality'' or evaluation ``noise'', our framework offers limited support for gauging the true influence of different evaluations. Further, the model doesn't yet support ranking coordination or cooperation risks based on their significance. Given that agents frequently make trade-offs driven by their network of dependency relationships, capturing the magnitude of coordination needs (e.g., how likely is it that an agent can predict the behaviour of another) and cooperation risks (e.g., which evaluations and incentives are an agent more likely to contribute effort towards) becomes important for devising robust mitigation strategies. Lastly, our model's scope is confined to a subset of coordination and cooperation risks, stemming primarily from specific dependency arrangements.

Notwithstanding its present limitations, the framework holds significant promise owing to its capacity to support the generation of actionable insights, as demonstrated through collaborative testing with the City.  Its impact potential impact has been demonstrated by its role in (1) facilitating the elucidation of integration challenges and (2) communicating the viability of alternative solutions.
The City's leadership was able to garner a deeper understanding of organization-wide challenges, which allowed them to develop informed strategies for managing the associated risks and promoting collaboration.
As we work towards integrating our framework into a versatile organizational decision-support tool, we envision users appending domain-specific axioms, expanding its relevance across diverse industries and settings.

Looking forward, we aim to enrich the representation of evaluations, incentives, and coordination needs, mitigating the aforementioned limitations. We also aim to model various organizational structures  (e.g.,\cite{2009_okhuysen.bechky_CoordinationOrganizationsIntegrativePerspective, 2010_gittell.seidner.ea_RelationalModelHowHighPerformanceWork}) and their influences on facilitating the level of knowledge and motivation necessary for mitigating coordination and cooperation risks.

\section{Conclusion}
In this paper, we have introduced a novel, semantics-based approach to empower organizations in identifying and addressing challenges related to coordination and cooperation. Motivated by the integration issues that surfaced in our ongoing collaboration with a major North American city,
we conducted an extensive review of the theoretical underpinnings of these phenomena in the context of organizational research, which yielded valuable insights into the nature of dependence and its relationship to organizational integration.
Existing enterprise modelling frameworks proved inadequate for accurately representing the intricate dependencies arising from real-world scenarios under consideration, prompting the development of a novel conceptual model that centers on operationalizing reward, epistemic, and outcome dependence, anchored in foundational constructs such as evaluations and incentives.
Our model's pragmatic utility was validated through its application to real-world scenarios. Its application proved instrumental in identifying and elucidating coordination and cooperation challenges, providing actionable insights for organizational decision-makers. 

We are confident in the transformative potential that our model holds for organizations, catalyzing enhanced coordination and cooperation. The early positive indicators from our collaboration with the City affirm this belief. While our work is relevant to organizational design,
we believe it has broader implications that extend beyond the field. 
The combination of semantic computing and organizational theory offers great promise for enhancing decision-making processes that involve coordinating and cooperating within organizations. By leveraging semantics to analyze and understand complex organizational dynamics, decision-makers gain a deeper appreciation of how various units and stakeholders interact and collaborate leading to better outcomes and greater effectiveness.

\small
\bibliographystyle{IEEEtranN}
\bibliography{ref.bib}

\end{document}